\documentclass{article}

\usepackage[preprint]{nips_2018}

\usepackage[utf8]{inputenc} 
\usepackage[T1]{fontenc}    

\usepackage{hyperref}       
\usepackage{url}            
\usepackage{booktabs}       
\usepackage{amsfonts}       
\usepackage{nicefrac}       
\usepackage{microtype}      

\usepackage{microtype}
\usepackage{graphicx}
\usepackage{wrapfig}
\usepackage{subfigure}
\usepackage{booktabs} 

\usepackage{amsmath,amssymb,amsfonts,wasysym,bm}
\DeclareMathOperator*{\expectation}{\mathbb{E}}
\DeclareMathOperator*{\argmin}{\rm argmin}

\usepackage{hyperref}

\usepackage{xspace}

\newcommand{\energy}{\mathcal{E}}

\title{Deep Energy Estimator Networks}

\author{Saeed Saremi$^\dagger$\\
Redwood Center for Theoretical Neuroscience\\
University of California, Berkeley \\
\texttt{saeed@berkeley.edu} \\
\And
Arash Mehrjou$^\dagger$ \\
Department of Empirical Inference \\
Max Planck Institute for Intelligent Systems\\
\texttt{arash.mehrjou@tuebingen.mpg.de} \\
\And
Bernhard Sch\"{o}lkopf \\
Department of Empirical Inference \\
Max Planck Institute for Intelligent Systems\\
\texttt{bs@tuebingen.mpg.de} \\
\And
Aapo Hyv\"{a}rinen \\
Gatsby Computational Neuroscience Unit \\
University College London, UK\\
\texttt{a.hyvarinen@ucl.ac.uk} \\
}

\begin{document}

\maketitle

\begin{abstract}
Density estimation is a fundamental problem in statistical learning. This problem is especially challenging for complex high-dimensional data due to the curse of dimensionality. A promising solution to this problem is given here in an inference-free hierarchical framework that is built on score matching. We revisit the Bayesian interpretation of the score function and the Parzen score matching, and construct a multilayer perceptron with a scalable objective for learning the  energy (i.e. the unnormalized log-density), which is then optimized with stochastic gradient descent. In addition, the resulting deep energy estimator network (DEEN) is designed as products of experts. We present the utility of DEEN in learning the energy, the score function, and in single-step denoising experiments for synthetic and high-dimensional data. We also diagnose stability problems in the direct estimation of the score function that had been observed for denoising autoencoders.
\end{abstract}

\section{Introduction}
\label{sec:Introduction}
{\it Universal density estimators\textemdash}  Learning the probability density of complex high-dimensional data is a challenging problem in machine learning. Our goal is to build a universal engine with a hierarchical structure to solve this problem robustly and efficiently.  This quest has a rich history, going back to the Boltzmann machine at the birth of the connectionist movement~\citep{hinton1986learning}, but an efficient, robust, and general-purpose method for continuous-valued data is still missing. 

{\it The curse of dimensionality\textemdash} The most fundamental road block for universal density estimation is the curse of dimensionality, the exponential growth of the ``volume'' of the data manifold with the dimension of the ambient space. In fact, this exponential growth is at the core of the shortcomings of kernel methods in density estimation. The curse can however be overcome by having strong priors on the data distribution. Perhaps, the strongest prior we currently have for natural distributions is that they are hierarchical and must be composed with deep architectures~\citep{bengio2009learning}. Guided by this prior, we demand the density estimation to be formulated in a deep hierarchical architecture. 

{\it The curse of inference\textemdash} One approach is to use latent variable models with hierarchical structure. The biggest challenge in training such models is that for complex distributions, the inference of the latent variables (i.e., over hidden units in the neural network) is intractable. While Markov chain Monte Carlo (MCMC) methods could be used, the problem is that natural distributions, for example natural images, have an intriguing property, also observed in physical systems near second-order phase transitions, that probability mass is concentrated in sharp ridges that are separated by large regions of low probability. This complex probability landscape is a major road block for learning; variational and MCMC methods have yet to meet the challenge~\citep{bengio2013representation}. An alternative is to use energy-based models, i.e.\ define an energy function which is essentially an unnormalized log-density function for the data. Again, however, such models tend to be intractable in the sense that the partition function (normalization constant) and its derivative cannot be computed easily --- their approximations by MCMC runs into the same problem as the inference of latent variables.

{\it Deterministic objectives\textemdash} These challenges facing MCMC methods gave rise to an important area of research in using {\it deterministic} objectives for parameter estimation of probabilistic models. One of the first examples of such frameworks for energy-based models is score matching~\citep{hyvarinen2005estimation}, which will be reviewed in a great depth below. A very early method was pseudolikelihood \citep{besag1974spatial}. More recently, minimum probability flow, with deep connections to score matching, was used to learn  Ising spin glasses~\citep{sohl2011new}. These methods define deterministic objective functions, although the optimization may be performed with stochastic methods.

{\it Our contributions\textemdash} In this work, with score matching as the foundation, we introduce a scalable and efficient algorithm to learn the energy of any data distribution in an inference-free hierarchical framework. We revisit the Bayesian estimation interpretation of the score function and Parzen score matching, and construct an objective, which is optimized  with stochastic gradient descent (SGD). The resulting Deep Energy Estimator Network (DEEN) is designed as products of experts (PoE), while the energy is efficiently estimated without resorting to MCMC computations and contrastive divergence approximations.  Our experiments were performed on two-dimensional synthetic data, MNIST, and the van Heteren data base of natural images. In addition, we diagnosed the problems with score function estimation in MLPs that bypass energy estimation; our framework avoids those problems as the score function is computed from the energy function itself.

\section{Background}
\label{sec:background}
{\it Energy-based models\textemdash} Assume we observe a random vector $x \in \mathbb{R}^d$ which has a probability density function (pdf) denoted by $p_x(.)$. We want to use a parameterized density model $p(x;\theta)$ with $\theta \in \mathbb{R}^N$, and estimate the parameter vector $\theta$ from observations of $x$; i.e. we want to approximate $p_x(.)$ by $p(.;\hat{\theta})$ for the estimated parameter value $\hat{\theta}$. In energy-based modelling, we start by defining a function 
$\mathcal{E}(x;\theta)$ which we use to define the model pdf as 
\begin{equation}
    p(x;\theta)=\frac{1}{Z(\theta)} \exp(-\mathcal{E}(x;\theta))
\end{equation}
Here we necessarily see the appearance of the partition function 
\begin{equation}
    Z(\theta)=\int \exp(-\mathcal{E}(x;\theta)) dx
\end{equation} 
which normalizes the pdf to have unit integral. In general, its computation is extremely difficult.

{\it Score matching\textemdash} Score matching estimates the parameters while completely avoiding the evaluation of the partition function $Z(\theta)$. The trick is that any multiplicative constant is eliminated in the score function $\psi: \mathbb{R}^d\rightarrow \mathbb{R}^d$ defined as \citep{hyvarinen2005estimation}:
\begin{equation}
	\psi(x;\theta) = \nabla_x \log p(x;\theta) = -\nabla_x \mathcal{E}(x;\theta).
\end{equation}
since the partition function does not depend on $x$.
\citeauthor{hyvarinen2005estimation} showed that under mild assumptions, the following score matching objective 
\begin{equation}
\label{eq:scorematching}
	\mathcal{L}(\theta) = \int_{x\in\mathbb{R}^d} p_x(x) \left \Vert \psi(x;\theta)-\psi_x(x)\right \Vert ^2 dx
\end{equation}
is (locally) {\it consistent}. In addition, he showed that we do not need to know the data score $\psi_x$ as integration by parts reduces $\mathcal{L}(\theta)$ to:
\begin{equation}
\label{eq:scorematching2}
	\mathcal{L}(\theta) = \int_{x\in\mathbb{R}^d} p_x(x) \left( \Vert \psi(x;\theta)\Vert^2 + 2~\nabla \cdot \psi(x;\theta) + \Vert \psi_x(x)\Vert^2\right) dx. 
\end{equation}
The term $\nabla \cdot \psi(x;\theta)$ is the divergence of the vector field $\psi$, equal to the negative trace of the Hessian of $\mathcal{E}$. The last term is not a function of $\theta$ and is dropped in the parameter estimation. For the finite dataset \begin{equation}\mathcal{D}  = \{ x^{(1)}, x^{(2)}, \cdots, x^{(n)}\},\end{equation} we obtain a finite-sample version by replacing $p_x$ by the empirical distribution:
\begin{equation}
		p_x(x) \approx {\sf E}(x) = \frac{1}{n} \sum_k \delta(x- x^{(k)}).
\end{equation}
So, in practice, $\theta$ was estimated by:
\begin{equation}
\label{eq:theta-hat}
	\hat{\theta} = \argmin_\theta \frac{1}{n} \sum_{k=1}^n  \Vert\psi(x^{(k)};\theta)\Vert^2  + 2~\nabla \cdot \psi(x^{(k)};\theta).
\end{equation}


{\it Kullback-Leibler vs. Fisher\textemdash} There is a formal link between maximum likelihood and score matching established by ~\cite{lyu2009interpretation}. By perturbing the observed data with the ``scale parameter'' $t$: $\xi = x + \sqrt{t}~\epsilon$, where $\epsilon \sim  {\sf Normal}(0,I)$, he proved  that the Fisher divergence is the derivative of the KL divergence with respect to the scale factor $t$. His interpretation of this result was that the score matching is more robust than maximum likelihood in parameter estimation as it searches for parameters that are relatively more stable to small noise perturbations in the training data.

{\it Score matching energy estimation\textemdash} The finite-sample score-matching objective of Eq.~\ref{eq:theta-hat} written in terms of the energy $\psi(x;\theta) = -\nabla_x \mathcal{E}(x;\theta)$ and expanded in the vector coordinates $i$ reduces to
\begin{equation}	
 \mathcal{L}_{0}(\theta) = \frac{1}{n} \sum_{i=1}^d \sum_{k=1}^n \left[\left( \frac{\partial \energy(x=x^{(k)};\theta)}{\partial x_i} \right)^2 - 2 \frac{\partial^2\energy(x=x^{(k)};\theta)}{\partial x_i^2}\right].
\end{equation}
\citep{kingma2010regularized} considered regularizing this objective where they arrived at regularized form of the score matching objective given by:
\begin{equation}
 \mathcal{L}_\lambda(\theta) = \mathcal{L}_{0}(\theta)+\lambda \sum_{i=1}^d \sum_{k=1}^n \frac{\partial^2\energy(x=x^{(k)};\theta)}{\partial x_i^2}.
\end{equation}
The second term above was obtained by a perturbative expansion of $\mathcal{L}_{0}(\theta)$ after adding a Gaussian cloud to the data points $x+\epsilon$ and ignoring the off-diagonal elements of the Hessian of the energy. \citeauthor{kingma2010regularized} further explored density estimation with the objective $ \mathcal{L}_\lambda(\theta)$  and parameterizing the energy $\mathcal{E}(x;\theta)$ with a MLP. However, the backpropagation of the full Hessian scales like $O(N^2)$, where $N$ is the total number of the parameters of the MLP. They found further approximations to the diagonal of the Hessian but the approximations were only valid for MLPs with a single hidden layer. Along these lines, approximation methods have been developed to estimate Hessians in neural networks~\citep{martens2012hessian}. We will see below a more principled way of regularizing score matching with Parzen score matching, and a stochastic version where the second-order estimations are avoided altogether. The objective we arrived at can also be motivated by the intriguing Bayesian interpretation of the score function that is discussed next.

{\it Bayesian interpretation\textemdash}  Score matching was originally motivated for its computational efficiency in the parameter estimation of unnormalized densities because of the key property that the partition function is eliminated  in the score function. However, there are also intriguing connections between score matching and a very different problem of Bayesian inference for signal restoration. This was established by~\citet{hyvarinen2008optimal} for small measurement noise, and generalized by~\citet{raphan2011least}. One way of looking at this is the Miyasawa denoising theorem: given noisy measurements $\xi$ of an underlying random variable $x$, when the measurement noise is ${\sf Normal}(0,\sigma^2 I)$, the Bayesian least-square estimator $\hat{x}(\xi) = \int x P(x|\xi) dx$ reduces to \begin{equation} \label{eq:nebls} \hat{x}(\xi) = \xi+\sigma^2 \psi(\xi;\theta),\end{equation} valid for {\it any level of noise $\sigma$}, with $\psi$ being the score function of the noisy data $\xi$. Thus, estimating the score function is essential for denoising. \citeauthor{raphan2011least} referred to this as {\it single-step denoising}, and generalized the result for a large class of measurement processes beyond Gaussian noise.

{\it Parzen density score matching\textemdash} ~\citet{vincent2011connection} studied score matching using a non-parametric estimator, replacing $p_x(.)$ by the Parzen windows density estimator $\sf P$:
\begin{equation} \label{eq:P}
	{\sf P}(\xi) = \frac{1}{n} \sum_k {\sf S}(\xi|x^{(k)}),
\end{equation}
where ${\sf S}$ is the smoothing kernel. The integration of Eq.~\ref{eq:scorematching} is then taken with respect to ${\sf P}(\cdot)$, and $\psi_x$ is replaced accordingly:
\begin{equation}
\label{eq:parzenscore}
	\mathcal{L}_{\sf parzen}(\theta) = \expectation_{\xi \sim \sf P} \left\Vert \psi(\xi;\theta)-\frac{\partial \log {\sf P}(\xi)}{\partial \xi}\right\Vert ^2.
\end{equation}
He then proved that the Parzen score matching objective is equivalent to the following ``denoising'' objective 
\begin{equation}
\label{eq:parzenvsdenoising}
\expectation_{(x,\xi)\sim{\sf J}} \left\Vert \psi(\xi;\theta)-\frac{\partial \log {\sf S}(\xi|x)}{\partial \xi} \right\Vert^2,
\end{equation}
with
\begin{equation}
\label{eq:jointdensity}
	{\sf J}(x,\xi) = {\sf S}(\xi|x) {\sf E}(x).
\end{equation}
for {\it any} smoothing kernel (see the appendix in~\citet{vincent2011connection} for the proof.)  The difficult kernel density estimation $\partial \log {\sf P}(\xi)/\partial \xi$ was thus replaced with $\partial \log {\sf S}(\xi|x)/\partial \xi$, which opens up the possibility of using a simpler stochastic implementation as we will see below.

{\it Single-step denoising interpretation\textemdash} Next we explain the denoising interpretation of the Parzen score objective $\mathcal{L}_{\sf parzen}(\theta)$ \citep{vincent2011connection}. This becomes clear by taking the Parzen kernel to be the conditional Gaussian density:
\begin{equation}
\label{eq:gausskernel}
	{\sf S}(\xi|x) = \frac{1}{(2\pi)^{d/2} \sigma^d} \exp\left( - \frac{\Vert \xi-x \Vert^2}{2\sigma^2}\right),
\end{equation}where after scaling by $\sigma^4$, the denoising objective in (\ref{eq:parzenvsdenoising}) reduces to:
\begin{equation}
\label{eq:vincent_denoising}
\mathcal{L}_{\sf denoising}(\theta) = \expectation_{(x,\xi)\sim{\sf J}} \left\Vert x-(\xi + \sigma^2 \psi(\xi;\theta)) \right\Vert^2 = \expectation_{(x,\xi)\sim{\sf J}} \left\Vert x-\hat{x}(\xi) \right\Vert^2.
\end{equation}
Note that from Eq.~\ref{eq:jointdensity}, $x$ is clamped to the training data, and $\xi$ can be interpreted as Gaussian ``noise'' added to the training data. Optimization of $\mathcal{L}_{\sf denoising}(\theta)$ learns a score function $\psi$ such that $\hat{x}(\xi)$ comes as close as possible to $x$, which is why it was dubbed as the denoising score matching objective by~\citeauthor{vincent2011connection}. That being said, we prefer here to interpret the objective defined in Eq.~\ref{eq:vincent_denoising} as Parzen score matching instead of denoising. This is more principled since in Parzen estimation there is a notion of optimality in smoothing the distribution: The optimal kernel width is likely to be non-zero and non-infinitesimal, irrespective of whether the measurements are noise-free or not. This is in contrast to denoising, where $\sigma$ is a free parameter, and its optimal value is a less meaningful quantity.
 
\section{Deep Energy Estimator Network}
\label{sec:deen}

{\it MLP parameterization of the energy\textemdash} Our goal is to build a general-purpose, efficient estimator $\mathcal{E}(x;\hat{\theta})$ that closely approximates the true energy $-\log p_x(\cdot)$ of any complex data distribution, up to an additive constant. We construct this energy estimator based on the score matching framework. For many real-life data distributions, the energy landscape is very complex with many canyons/ridges of low-energy/high-density being separated by large flat regions of high-energy/low-density. Deep neural networks are natural candidates for shaping such complex energy landscapes and thus beating the curse of dimensionality. In fact, to obtain a model of great generality, we can leverage the universal approximation capacity of neural networks and use a neural network as the model of the energy (log-density):
\begin{equation}
    \mathcal{E}(x;\theta)={\sf MLP}(x;\theta)
\end{equation}
where ${\sf MLP}(x;\theta)$ means the output of a multi-layer perceptron with input $x$ and (weight) parameters $\theta$; it takes the inputs in $\mathbb{R}^d$ and outputs a scalar in $\mathbb{R}$.

{\it Scalable score matching\textemdash} One way to find an estimate $\theta$ for the energy $\mathcal{E}(x;\theta)$ would be to simply apply the basic definition of score matching. However, as discussed above, such an approach is often considered to be computationally prohibitive due to the appearance of second-order derivatives in the objective function~\citep{kingma2010regularized} and one must resort to approximations~\citep{martens2012estimating}. Our key insight here is to build on the Bayesian interpretation of the score function of \citeauthor{raphan2011least} and the Parzen score matching of \citeauthor{vincent2011connection}. This leads to a method which scales well to high dimensions, as well as to large datasets. 

{\it The DEEN objective\textemdash} Thus,  we use a multilayer perceptron that parametrizes $\mathcal{E}(x;\theta)$ and the score is given by  $\psi(x;\theta) = -\partial \mathcal{E}(x;\theta)/\partial x$. 
In this work we use the Gaussian kernel, which gives, based on Eq.~\ref{eq:parzenvsdenoising}, the objective
\begin{equation}
    \label{eq:deen}
	\mathcal{L}_{\sf DEEN}(\theta) = \expectation_{(x,\xi)\sim{\sf J}} \left\Vert x-\xi + \sigma^2\frac{\partial \mathcal{E}(\xi;\theta)}{\partial \xi } \right\Vert^2,
\end{equation}
For the MLP parameterization of the energy, the term $\partial \mathcal{E}(\xi;\theta)/\partial \xi $ is an acyclic directed computation graph mapping $\mathbb{R}^d  \rightarrow \mathbb{R}^d$, which can be built with automatic differentiation. The variance $\sigma^2$ is the smoothing hyperparameter of the Parzen kernel, which we choose by optimizing the test likelihood of the Parzen density over a validation set in our experiments. The objective is then optimized:
\begin{equation}
	\hat{\theta} = \argmin_\theta \mathcal{L}_{\sf DEEN}(\theta).
\end{equation}
{\it Hyperparameter $\sigma$\textemdash} Parzen score matching gave us an objective for energy estimation that did not involve estimating the Hessian of the energy. In addition, there is also an automatic regularization in DEEN because of the hyperparameter $\sigma$ of the Parzen smoothed density. Note that $\sigma$ is dataset dependent: as the dataset size $n$ increases, $\sigma$  decreases. Also, $\sigma$ is smaller for datasets with lower complexity (measured by entropy). However, the theoretical limit $\sigma\rightarrow 0$ is never within reach for complex datasets due to the curse of dimensionality as $n$ has to grow exponentially in dimension $d$.

{\it SGD optimization of $\mathcal{L}_{\sf DEEN}$\textemdash} For the dataset $\mathcal{D}  = \{ x^{(1)}, x^{(2)}, \cdots, x^{(n)}\}$, $\mathcal{L}_{\sf DEEN}$ is approximated by constructing the joint set $\mathcal{J}$:
\begin{eqnarray}
	\mathcal{J} &=& \{ (x^{(1)},\mathcal{X}^{(1)}), (x^{(2)},\mathcal{X}^{(2)}), \cdots, (x^{(n)},\mathcal{X}^{(n)})\}, \\
	\mathcal{X}^{(i)} &=& \{ \xi^{(i,1)},\xi^{(i,2)},\dots,\xi^{(i,m)}\},
\end{eqnarray}
   
where $\xi^{(i,j)}$ are iid samples from the smoothing Kernel ${\sf S}(\xi|x^{(i)})$, i.e.\ data with noise added. The objective is then approximated by an empirical version:
\begin{equation}
\label{eq:deenapprox}
    \mathcal{L}_{\sf DEEN}(\theta) \approx \sum_{i=1}^n \sum_{j=1}^m \left\Vert x^{(i)}-\xi^{(i,j)}+\sigma^2\frac{\partial \mathcal{E}(\xi=\xi^{(i,j)};\theta)}{\partial \xi}
 \right\Vert^2
  \end{equation}
After constructing the set with $m\gg1$ (for a better approximation), we would arrive at a ``deterministic'' objective, which is optimized with SGD with two sources of stochasticity: the random choice of mini batches $x^{(i)}$ and $\mathcal{X}^{(i)}$. In our double-SGD implementation we set the mini batch size of the second set $\mathcal{X}^{(i)}$ to be one, i.e. a single noisy observation $\xi$ for each data point. We emphasize that the computations are scalable, unlike a straight-forward application of the original score matching.

{\it DEEN as PoE\textemdash} In our MLP architecture, the energy $\mathcal{E}(x;\theta=\{w^{(1)},w^{(2)},\cdots,w^{(L+1)}\})$ is constructed linearly from the last hidden layer $h^{(L)}$:\begin{equation} \mathcal{E}(x;\theta) = \sum_{\alpha} w^{(L+1)}_\alpha h^{(L)}_\alpha(x;\{w^{(1)},w^{(2)},\cdots,w^{(L)}\}) = \sum_\alpha \varepsilon_\alpha(x;\theta), \end{equation} where the sums are over the hidden units in the last layer. The factorization $p(x;\theta) \propto \prod_\alpha \exp(-\varepsilon_\alpha(x;\theta))$ corresponds to the products of experts (PoE) where $\varepsilon_\alpha$ are the experts parametrized by a neural network. PoE is powerful in partitioning the space in a distributed representation formed by $\varepsilon_\alpha(x;\theta)$, in contrast to mixture models that partition the space with one region per expert~\citep{hinton1999products}. Despite PoE's immense potentials in beating the curse of dimensionality, the estimation of the derivative of the partition function $Z(\theta)$ plagues the learning, and approximations like the contrastive divergence~\citep{hinton2006training,hinton2006unsupervised} has yet to overcome the curse of inference for  complex distributions. In DEEN, powered by score matching, no such approximations are made. In this work, all the experts share parameters, but our framework contains ``mixtures of DEENs'', where the energy $\mathcal{E}(x;\Theta = \{\theta^1,\theta^2,\cdots,\theta^M\})$ is constructed by averaging the outputs of $M$ neural networks if we loosely consider each hidden unit as a neural network in itself. Although a single DEEN is by itself a universal estimator, but imposing structures by breaking the neural network into sub-networks would help us better understand the experts specilizations, which presumably will be crucial for representation learning with the emergent PoE.

{\it DEEN and denoising autoencoders\textemdash} An important result in the score matching literature is due to~\citet{alain2014regularized}, who discovered a strong and surprising link between  regularized (contractive/denoising) autoencoders and the score function itself. Denoting $r^*_{\sigma}(x)$ as the {\it optimal reconstruction} of the $\sigma$-regularized autoencoder, they arrived at the result:\begin{equation} \lim_{\sigma\rightarrow 0} (r^*_{\sigma}(x)-x)/\sigma^2 = \nabla_x \log p(x),\end{equation} which is similar to the single-step denoising interpretation of DEEN on the surface. But DEEN {\it is not an autoencoder}, it learns the energy and thus the score function {\it without a decoder}, and as we emphasized, the optimal $\sigma$ in the kernel density estimator {\it will not be small} for any complex dataset. In addition, denoising autoencoders have stability issues in learning the score function, which we will elaborate further shortly. We think DEEN is better regularized in that respect, since it has no constraint on $\sigma$ which is a {\it hyperparameter} (dataset dependent). In fact, a natural estimator for $\sigma$ is obtained by optimizing the test likelihood of the Parzen density over a validation set.
 
{\it Energy estimation vs score estimation\textemdash}  What if, alternatively, one is only interested in estimating the score function and not the energy? A multilayer perceptron $\psi(x;\theta) = {\sf MLP}(x;\theta)$ that maps $\mathbb{R}^d$ to $\mathbb{R}^d$ --- referred here as deep score matching (DSM) --- can be trained to learn the score  by optimizing the  objective  given in Eq.~\ref{eq:vincent_denoising}. However, the ``direct'' score function estimation with MLPs are not robust --- this problem was first observed in ~\citet{alain2014regularized} for denoising autoencoders (see their Fig. 7). We provide a detailed diagnosis of this problem in our 2D experiments below. DEEN does not suffer from such stability issues of the score function estimation by construction.

%

\section{Experiments}
\label{sec:experiments}

\begin{figure}[b!]%
\begin{center}
    \begin{subfigure}[$q(x;\theta_t)$, $ t=1e3$]{\includegraphics[width=0.23\textwidth]{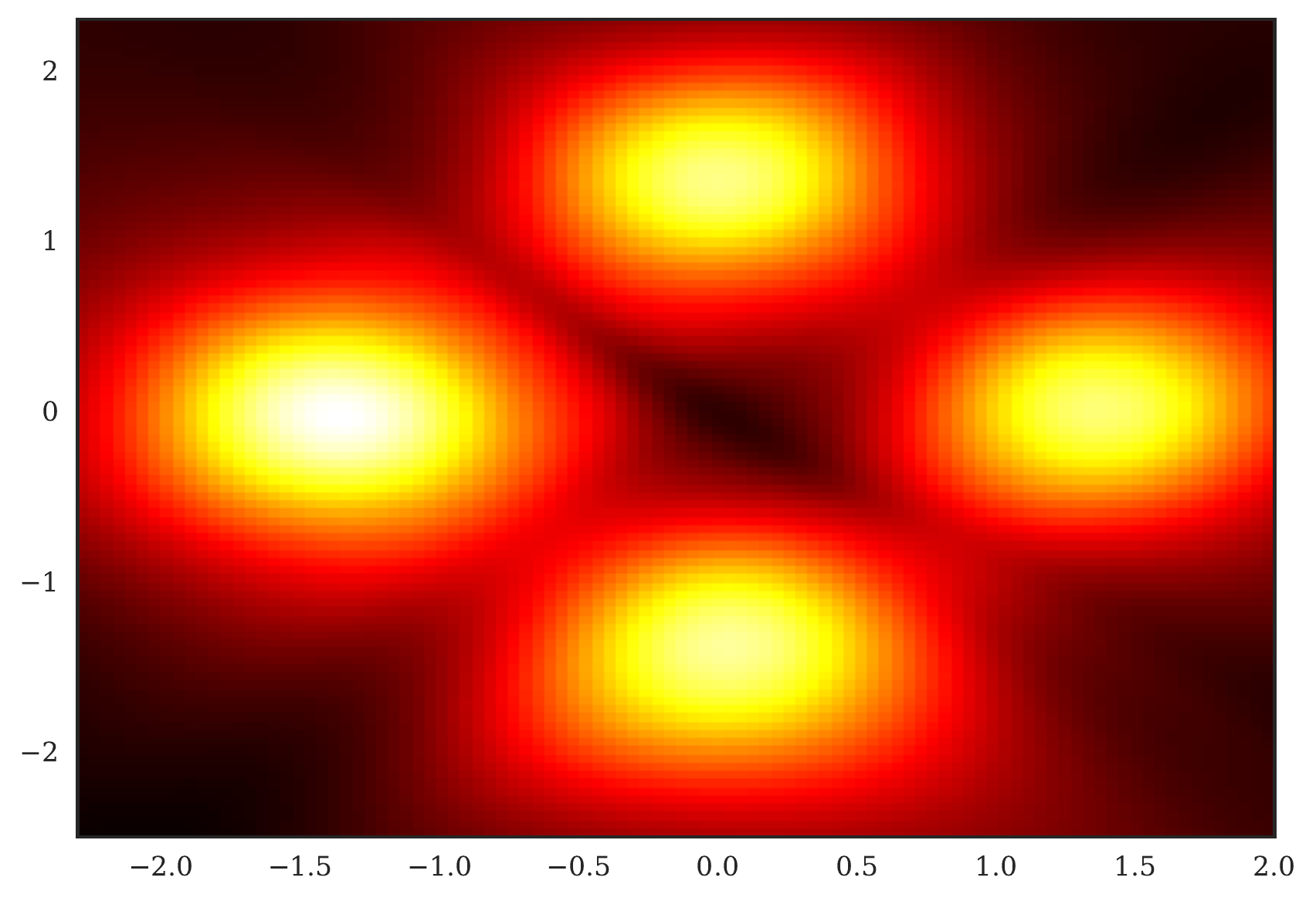} }%
    \end{subfigure}
    \begin{subfigure}[$-\nabla_x \mathcal{E}(x;\theta_t)$]{\includegraphics[width=0.23\textwidth]{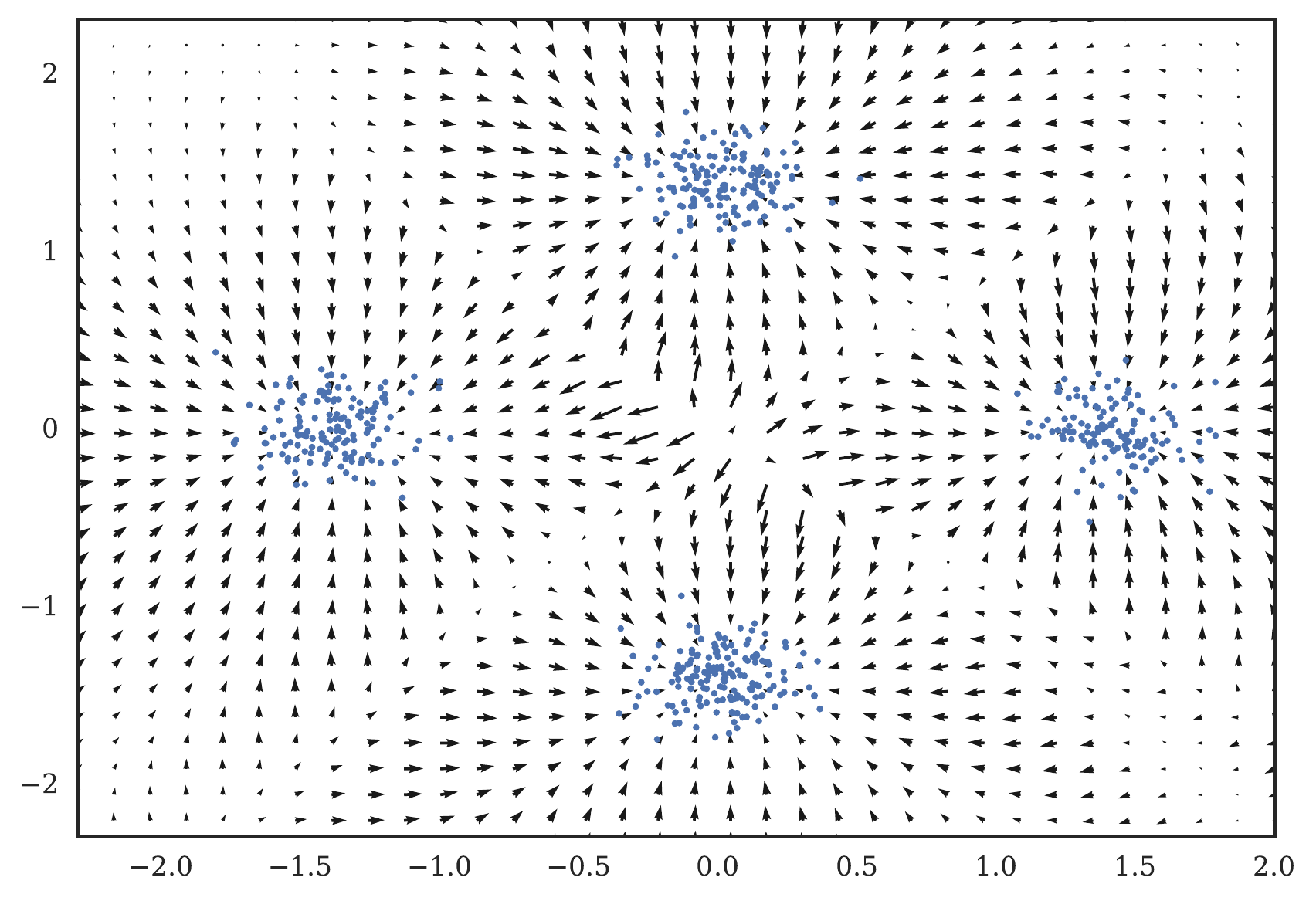} }%
    \end{subfigure}
        \begin{subfigure}[$q(x;\theta_t)$, $ t=1e3$]{\includegraphics[width=0.23\textwidth]{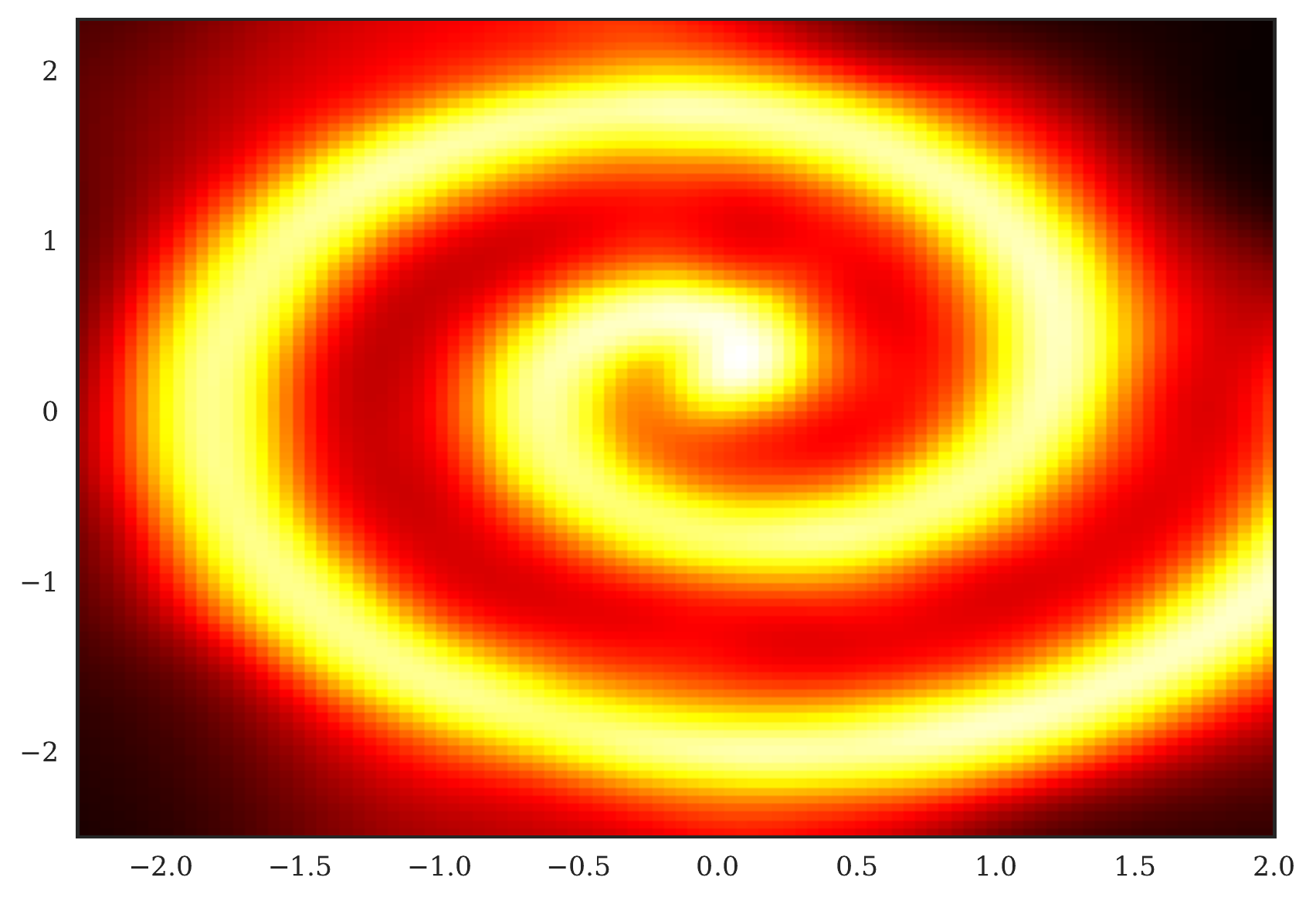} }%
    \end{subfigure}
    \begin{subfigure}[$-\nabla_x \mathcal{E}(x;\theta_t)$]{\includegraphics[width=0.23\textwidth]{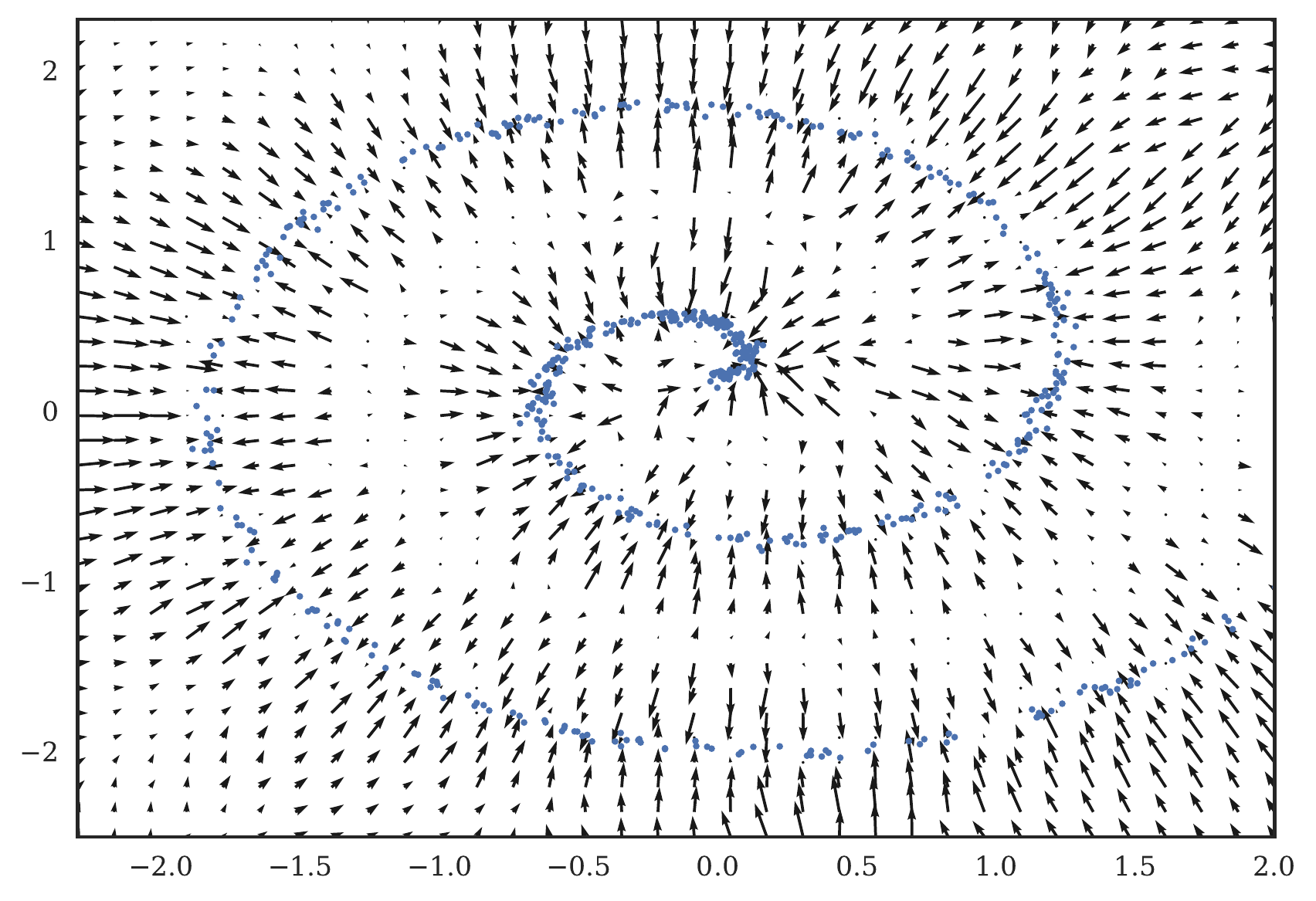} }%
    \end{subfigure}

    \begin{subfigure}
    {\includegraphics[width=0.49\textwidth]{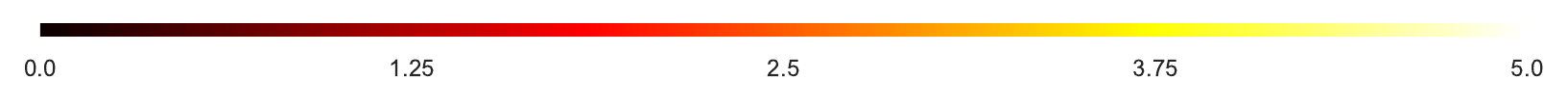} }%
    \end{subfigure}
    \begin{subfigure}
    {\includegraphics[width=0.49\textwidth]{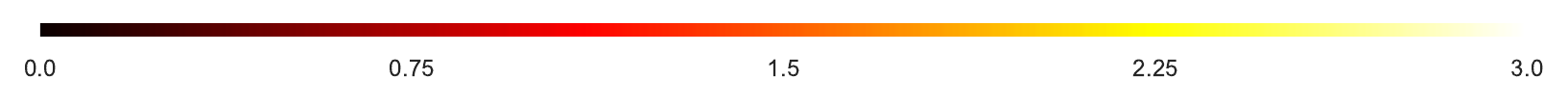} }%
    \end{subfigure}\\

    \begin{subfigure}[ $\psi(x;\theta_t)$,  $t= 1e4$]{\includegraphics[width=0.23\textwidth]{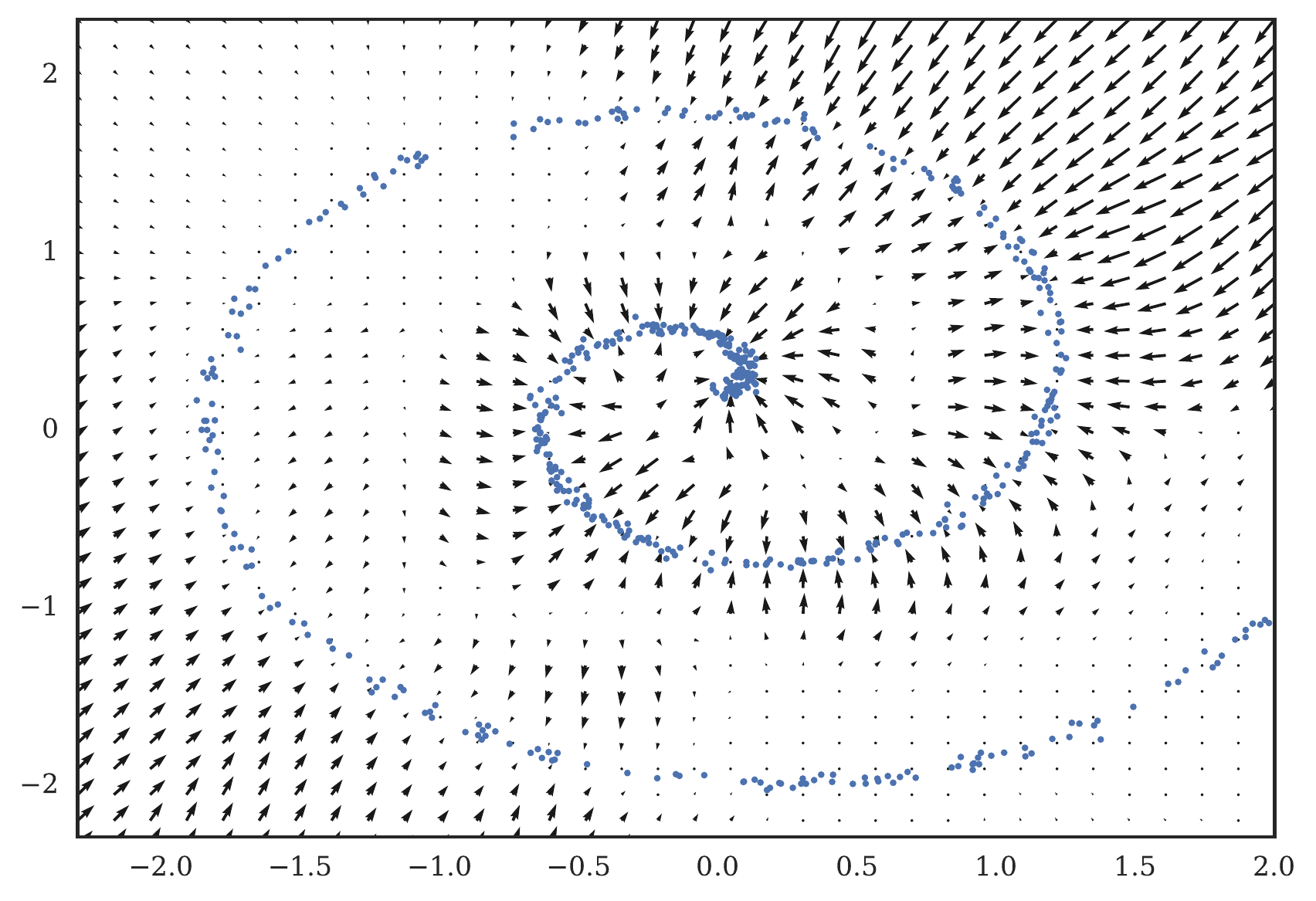} }%
    \end{subfigure}
    \begin{subfigure}[$|\nabla \times\psi(x;\theta_t)|$]{\includegraphics[width=0.23\textwidth]{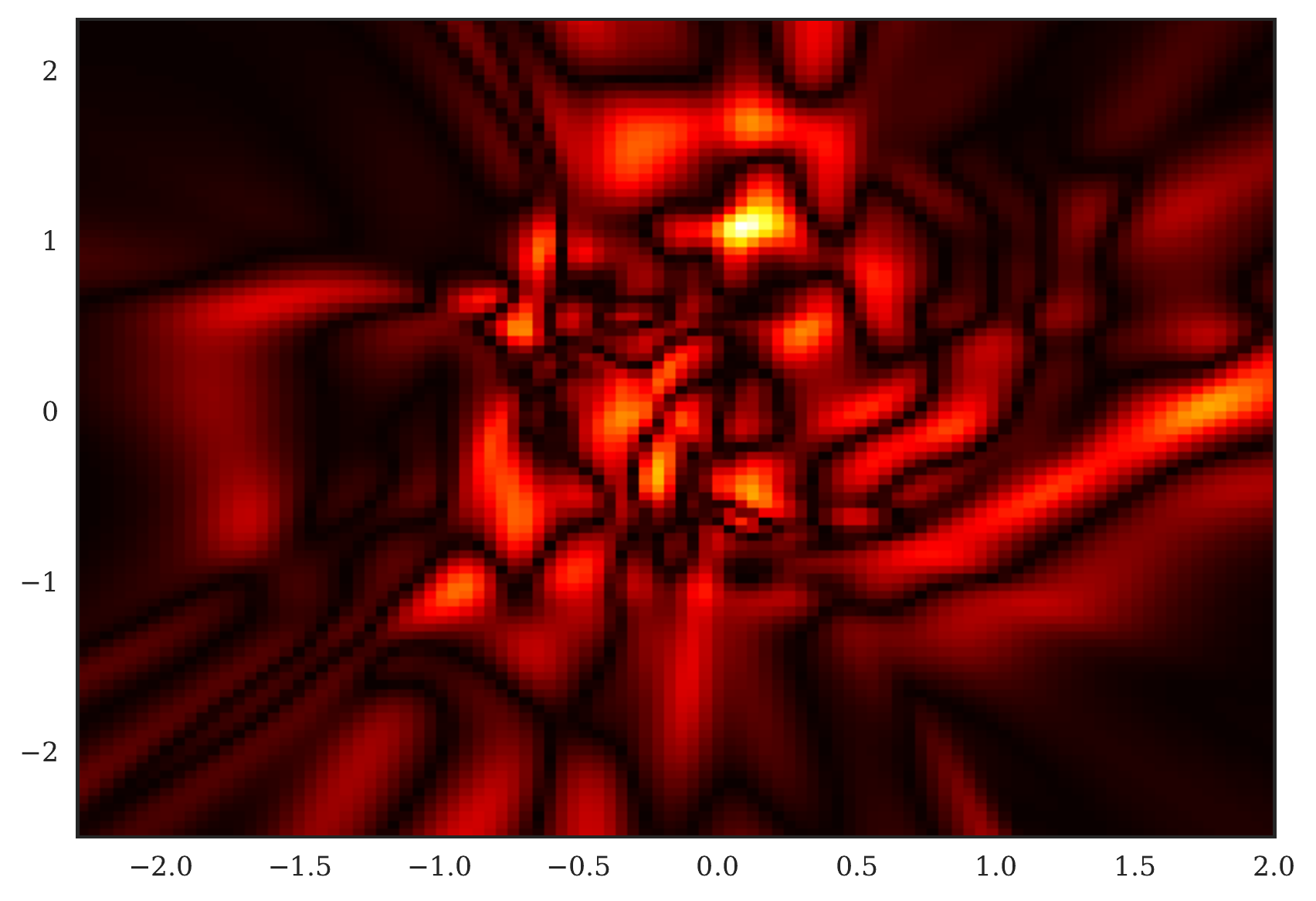} }%
    \end{subfigure}
    \begin{subfigure}[MoG, $\sigma=0.1$]{\includegraphics[width=0.23\textwidth]{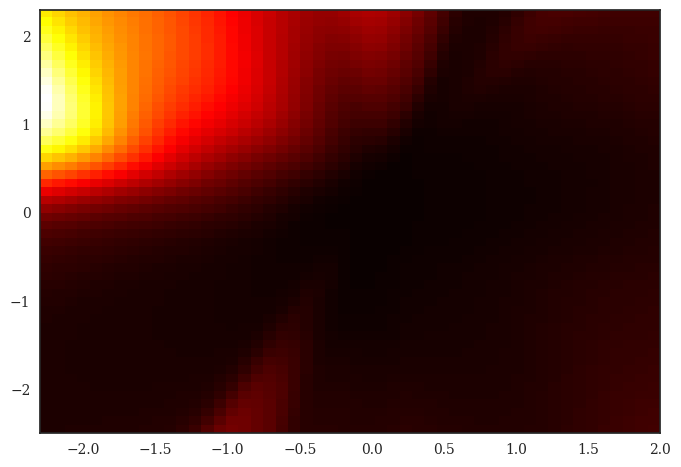} }%
    \end{subfigure}
    \begin{subfigure}[Spiral, $\sigma=0.1$]{\includegraphics[width=0.23\textwidth]{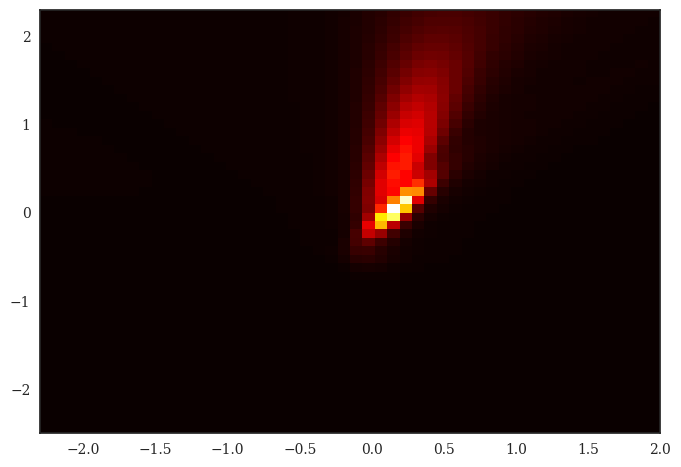} }%
    \end{subfigure}\\
        \begin{subfigure}
    {\includegraphics[width=0.49\textwidth]{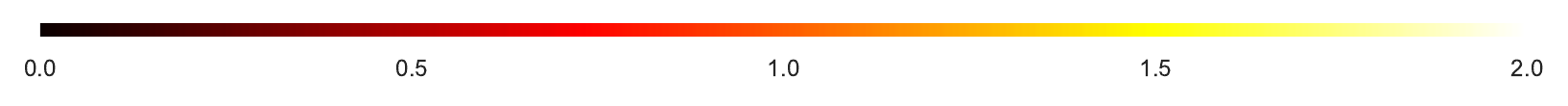}}
    \end{subfigure}
        \begin{subfigure}
    {\includegraphics[width=0.49\textwidth]{nipsfigs/heatbar_curl.pdf}}
    \end{subfigure}

    \caption{{\it 2D experiments\textemdash} {\it (a)(c)} DEEN's performance after 1000 SGD iterations in learning the energy $\mathcal{E}$, where $q(x;\theta_t) = \exp(-\mathcal{E}(x;\theta_t))$ is plotted. {\it (b)(d)} The score function $\psi = -\nabla \mathcal{E}$ {\it computed} from DEEN is plotted. {\it (g)(h)} DSM's {\it failure}  (akin to instabilities of denoising autoencoders in learning the score function) in learning the right score diagnosed by $\nabla \times\psi$ which {\it must be zero.} {\it (i)(j)} Two examples of the failure modes of contrastive divergence explained at the end of Experiments Section for mixtures of Gaussians and the spiral. Colorbars are below each subfigure.
    }
    \label{fig:2d-experiments}
\end{center}
\end{figure}

{\it Two-dimensional experiments\textemdash} We estimated the energy function of a spiral toy data set as well as a Gaussian mixture data set presented in Fig.~\ref{fig:2d-experiments}{\it (a)--(d)}. The MLP architecture is fully connected with three hidden layers of tanh nonlinearities, and a linear last layer: $x\rightarrow h^{(1)} \rightarrow h^{(2)} \rightarrow h^{(3)}  \rightarrow \mathcal{E}$, $\dim(h^{(l)})=32$. DEEN estimated the energy accurately with well-behaved training curves.

{\it Failures of estimating the score function directly\textemdash}  We experimented with learning the score function directly in our DSM network discussed at the end of Sec.~\ref{sec:deen}. Whether the learned vector field can be written as a gradient of a scalar field can be diagnosed and visualized in 2D by using the curl operator: $(\nabla \times \psi)_3 =  \partial_1 \psi_2-\partial_2 \psi_1$. The identity $\nabla \times \psi(x)=0$ is a necessary and sufficient condition for the existence of a scalar field $\mathcal{E}(x)$ such that $\psi(x) = - \nabla\mathcal{E}(x)$: checking $\nabla \times \psi(x)=0$ is therefore an easy sanity check that the learned score must pass. But, in practice, training the MLP with the DSM objective learns a vector field such that $\nabla \times \psi(x;\hat{\theta}) \neq 0$, far from zero in fact as we show in our experiments in Fig.~\ref{fig:2d-experiments} {\it(g)(h)}. Note that DEEN is immune to this type of instability, as the score function is {\it computed} from the learned energy function itself.

\begin{figure}[t!]%
\begin{center}
	\begin{subfigure}[$\langle \mathcal{L}_{\sf DEEN}(\theta_t;\sigma=0.14)\rangle$ vs. SGD $t$]{\includegraphics[width=0.36\textwidth]{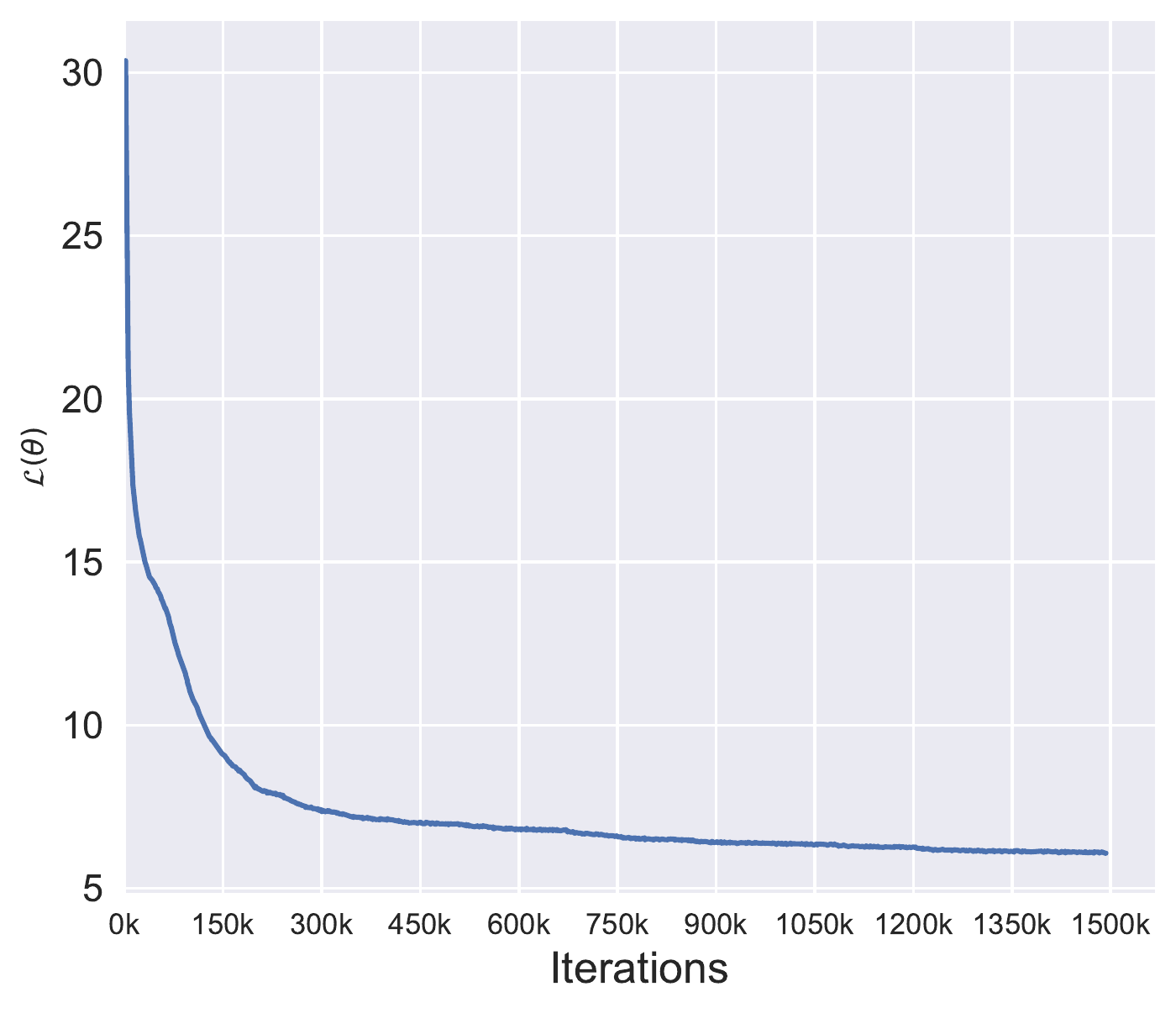} }%
	\end{subfigure} 
    \begin{subfigure}[ additive noise, $\sigma'=0.3$]{\includegraphics[width=0.3\textwidth]{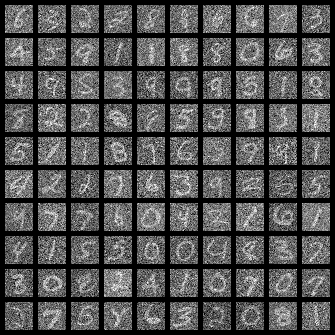} }%
    \end{subfigure}
    \begin{subfigure}[{\it single-step}	 denoising]{\includegraphics[width=0.3\textwidth]{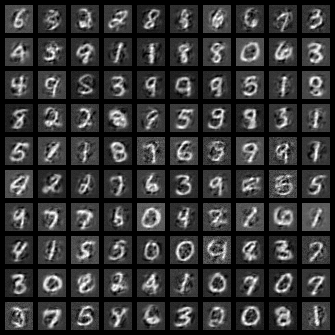} }%
    \end{subfigure}
    \caption{{\it DEEN on MNIST\textemdash} {\it (a)} The running average of $\mathcal{L}_{\sf DEEN}$, with the hyperparamater $\sigma=0.14$.  {\it (b)(c)} The learned energy function $\mathcal{E}(x;\theta_\infty)$ is used for the single-step denoising with $\sigma' \neq \sigma$. 
    }
    \label{fig:mnist}
\end{center}
\end{figure}

{\it DEEN on MNIST\textemdash} In MNIST experiments, the hyperparameter $\sigma=0.14$ was first chosen by maximizing the test likelihood of the Parzen estimator. Fig.~\ref{fig:mnist}{\it(a)} shows the running average of $\mathcal{L}_{\sf DEEN}$ vs. SGD iterations for $\sigma=0.14$. We tested DEEN on single-step denoising (SSD) experiments, presented in Fig.~\ref{fig:mnist} {\it(b)(c)}. Theoretically,  SSD should be done with the same value of $\sigma$, but we also tested it with different values, presented here are the experiments with $\sigma'=0.3$.  The MLP architecture is the same as 2D experiments but with $\dim(h^{(1,2)})=128, \dim(h^{(3)})=256$.

\begin{figure}[t!]%
 \begin{center}
     \begin{subfigure}[noisy, err/pixel = 0.0308]{\includegraphics[width=0.32\textwidth]{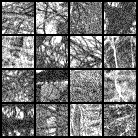} }%
     \end{subfigure}
     \begin{subfigure}[MG baseline, err/pixel = 0.0174]{\includegraphics[width=0.32\textwidth]{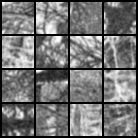} }%
     \end{subfigure}
     \begin{subfigure}[DEEN, err/pixel = 0.0106]{\includegraphics[width=0.32\textwidth]{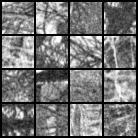} }%
     \end{subfigure}
     \caption{{\it DEEN on natural images\textemdash} {\it (a)} images in test set with patch-dependent noise level, explained in the Experiments Section. {\it (b)} denoising performance with median + Gaussian filter, standard in signal processing. err/pixel is reported here. {\it (c)} single-step denoising with DEEN. }
     \label{fig:vanhat}
 \end{center}
 \end{figure}

{\it DEEN on natural images\textemdash}  We tested DEEN's performance on $32\times32$ patches extracted from the van Hateren database of natural images~\citep{van1998independent}. This time, after training, we also tested the denoising for additive patch-dependent noise $0.5\times {\rm std}(x)\times \nu$ where $\nu \sim {\sf Normal}(0,I)$. Fig.~\ref{fig:vanhat}{\it(a)} shows examples of the noisy patches in the test set.
Fig.~\ref{fig:vanhat}{\it(b)} is the denoising by median + Gaussian filters, a baseline in signal processing~\citep{gonzalez2012digital}. Fig.~\ref{fig:vanhat} {\it(c)} is the single-step denoising from the energy function. The err/pixel reported here is equal to $\Vert x-\hat{x} \Vert^2$ normalized by the number of pixels. The MLP architecture here is again fully connected, but with five hidden layers of tanh nonlinearities, and  $\dim(h^{(1,2,3)})=256, \dim(h^{(4,5)})=512$. All experiments presented so far were implemented in Tensorflow~\citep{abadi2016tensorflow} using Adam~\citep{kingma2014adam} as the optimizer with the learning rate of 0.001. 

 {\it CD failures\textemdash} We also experimented with an alternative formulation of score matching by means of contrastive divergence (CD) with Langevin updates~\citep{hyvarinen2007connections}. However, the equivalence of Langevin dynamics CD and score matching is only exact in asymptotic regimes, and in our 2D experiments the CD updates showed severe mode collapse that we report here. We trained the same MLP architecture as the DEEN experiments in 2D using the CD updates with negative samples:\begin{eqnarray}
 	 x^- &=& x - \sigma^2 \nabla_x \mathcal{E}(x;\theta_t) + \sqrt{2} \sigma \nu \\
     \theta_{t+1} &=& \theta_t +\epsilon[ \nabla_\theta \mathcal{E}(x^-; \theta_t) - \nabla_\theta \mathcal{E}(x;\theta_t) ]
    \label{eq:cd_loss}
\end{eqnarray}
where $\nu \sim {\sf Normal}(0,I)$, and $\epsilon$ is the learning rate set to $0.001$. The above dynamics does not optimize a well-defined objective, and in our experiments, we observed severe mode collapse in energy estimation for both mixtures of Gaussians and spiral. Examples of the failures are reported in Fig.~\ref{fig:2d-experiments} {\it (i)(j)}. We did some experimentations with the hyperparameter $\sigma$, but the experiments shown here were for the same value of $\sigma=0.1$ as in the other 2D experiments in that figure.

\section{Discussion}

{\it Our main contribution\textemdash} We presented a hierarchical, yet efficient solution for approximating the energy of data distributions. This was done in terms of a multilayer perceptron, thus allowing universal approximation. The objective is principled, rooted in score matching with consistent estimation guarantees, and the training has the complexity of backpropagation. 
We further diagnosed instabilities of the denoising autoencoders in learning the score function. DEEN is immune to those particular instabilities since the score function is directly computed from the learned energy function.

{\it Scalable unsupervised learning algorithms\textemdash} Our approach fits within the general movement, prominent in recent years, which employs the SGD's ``unreasonable effectiveness'' in training deep neural networks at the service of unsupervised learning. 
An important step is to formulate a learning objective that allows automatic differentiation and some variant of backpropagation to work with linear complexity in the number of dimensions and parameters. With score matching, we had to resort to the Bayesian interpretation and the Parzen kernel formulation of score matching to achieve this.

{\it Different aspects of unsupervised learning\textemdash} It is important to notice that unsupervised learning is not a single well-defined problem; in fact there is a number of different goals, and different methods excel in different goals (see for example \citet{theis2015note} for discussions on generative models.)  One example is the problem of {\it generative modeling} and generative adversarial networks are designed for this task~\citep{goodfellow2014generative}. Another goal is {\it posterior inference} for hierarchical graphical models. Variational autoencoders were designed to tackle this problem with a variational objective~\citep{kingma2013auto}. Another problem is the problem of disentangling factors of variation, i.e. the problem of {\it nonlinear ICA}. This problem was recently reformulated for time series, solving the nonidentifiability problem~\citep{hyvarinen2016unsupervised}. In our work here, we looked at another important problem in unsupervised learning, {\it density estimation} of unnormalized densities. The score matching objective $\mathcal{L}_{\sf DEEN}$ which is the foundation for our work is surprisingly simple, but its power is displayed when the energy $\mathcal{E}(\cdot;\theta)$ is parameterized by a deep neural network.

{\it Future directions\textemdash} Two important applications of energy estimators include semi-supervised learning and generative modeling. Presumably, the network has learned a useful representation of the data in the last hidden layer(s), and such a hidden representation would enable very efficient supervised learning. In that regard, understanding the PoE underlying DEEN is an important next step. In addition, the energy function could be directly plugged-in in various MCMC methods. The second problem is perhaps harder since Hamiltonian Monte Carlo (HMC) methods are typically quite slow. But there are important new developments on the HMC front~\citep{girolami2011riemann,levy2017generalizing}, which DEEN can build on. Among closely related work to our present work, we should mention \textit{noise-contrastive estimation (NCE)}, which also has the potential for estimating the energy function using general function approximation by deep learning \citep{gutmann2012noise}. A further important avenue would be to see if NCE provides energy estimates of the same quality as score matching; presumably, an intelligent way of choosing the ``noise'' distribution may be important.

{\it Stein contrastive divergence\textemdash} In preparing this submission, we came across the work by ~\citet{liu2017learning} and the Stein contrastive divergence introduced there for learning energy models. There are deep connections between score matching and their approach, which was emphasized in their paper, where at the same time they also avoided second-order derivatives. However, Stein contrastive divergence is fundamentally based on infinitesimal negative-sample generations and CD-type updates (Eq. 13 and Algorithm 1 in the reference). DEEN by contrast is fundamentally based on backpropagation and SGD on an objective. It is important to understand the possible links between the two frameworks and compare the performance and the efficiency of the two algorithms.

\bibliography{scorematching}

\end{document}